\documentclass[10pt,twocolumn,letterpaper]{article}

 \usepackage[pagenumbers]{cvpr} % To force page numbers, e.g. for an arXiv version
\usepackage{algpseudocode}
\usepackage{amsmath}
\usepackage{tcolorbox}
\usepackage{algorithm}

\definecolor{cvprblue}{rgb}{0.21,0.49,0.74}
\usepackage[pagebackref,breaklinks,colorlinks,citecolor=cvprblue]{hyperref}

\usepackage{geometry}

\newtheorem{definition}{Definition}

\usepackage{xcolor,colortbl}
\definecolor{green}{rgb}{0.1,0.1,0.1}
\definecolor{tabfirst}{rgb}{1, 0.7, 0.7}
\definecolor{tabsecond}{rgb}{1, 0.85, 0.7} 
\definecolor{tabthird}{rgb}{1, 1, 0.7} 
\usepackage{hyperref}

\def\paperID{*****} % *** Enter the Paper ID here

\title{Reinforcement Learning from Diffusion Feedback: \textit{\space\space\space\space\space\space\space\space\space\space\space\space\space\space\space\space\space\space\space\space\space\space\space\space\space\space\space\space\space\space\space\space\space\space\space\space\space\space\space\space\space\space\space\space Q* for Image Search}}

\author{Aboli Marathe\\
Machine Learning Department\\
Carnegie Mellon University\\
{\tt\small abolim@cs.cmu.edu}\\
}

\begin{document}

\twocolumn[{
\renewcommand\twocolumn[1][]{#1}
\maketitle
\begin{center}
    \centering
    \vspace*{-.8cm}
    \includegraphics[width=0.8\textwidth]{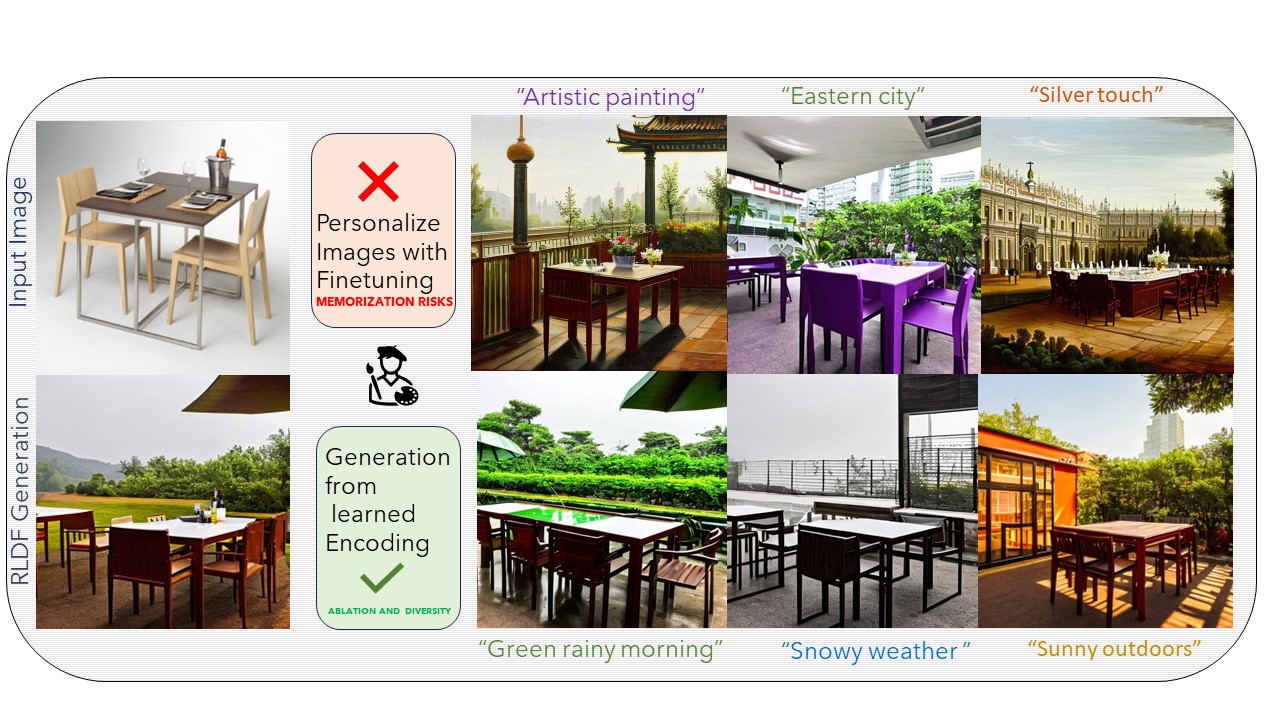}
    \vspace*{-.6cm}
    \captionof{figure}{With just a single input image (left), \emph{RLDF} can generate diverse semantically similar images (bottom and right). RLDF requires no text guidance or fine-tuning and can be personalized by simple actions in the the semantic encoding space.}
\label{fig1}
\end{center}
}]

\begin{abstract}

Large vision-language models are steadily gaining personalization capabilities at the cost of fine-tuning or data augmentation. We present two models for image generation using model-agnostic learning that align semantic priors with generative capabilities. RLDF, or Reinforcement Learning from Diffusion Feedback, is a singular approach for visual imitation through prior-preserving reward function guidance. This employs Q-learning (with standard Q*) for generation and follows a semantic-rewarded trajectory for image search through finite encoding-tailored actions. The second proposed method, noisy diffusion gradient, is optimization driven. At the root of both methods is a special CFG encoding that we propose for continual semantic guidance. Using only a single input image and no text input, RLDF generates high-quality images over varied domains including retail, sports and agriculture showcasing class-consistency and strong visual diversity. Project website: \href{https://infernolia.github.io/RLDF}{https://infernolia.github.io/RLDF}.
\end{abstract}
 \vspace*{-0.5cm}
\section{Introduction}
\vspace*{-.21cm}

Recent breakthroughs in text-to-image models were quickly followed by the adoption of human intervention in visual prompt engineering. Humans typically use a visual feedback loop to prompt VLMs and modify the prompts until they arrive at the desired images — \textit{an afternoon of merriment for some, yet an expensive ordeal for another}. What if we could align generative models with proxies for human perception using semantic diffusion feedback?

We aim to single-handedly eliminate this bottleneck of human feedback by context-driven image generation guided by semantic priors. Critically, we attempt to avoid subject-driven generation to mitigate propagation of memorized examples and instead focus on the class of subject-driven generations. The task can be summarized as \textbf{zero-shot semantic-guided generation for imitation using visual-prompting of VLMs}. To simulate the long tail, we impute bias-equalizing corrections into the generative process to obtain diversity in generated examples. 

\paragraph{}
Our  contributions are many-fold. 
 \begin{enumerate}

    \item We propose RLDF and nDg models for class-driven semantic imitation using only a single image input. We test these models on multiple domains (Figure \ref{fig14}) and on a full ImageNet clone  (Figure \ref{fig15}), for evaluating usability on real-world benchmarks.
    \item We demonstrate highly effective model-agnostic stability across DALLE-2, SD 1.4 and SD 2.1 models in a plug-and-play mechanism with convergence guarantees. 
    \item RLDF demonstrates generalization across object and action spaces, allowing extensive personalization capabilities.
\item The models implicitly attempt to ablate training concepts which can assist for copyright protection and style removal.

\end{enumerate}

\begin{figure}
\includegraphics[width = 0.98\linewidth]{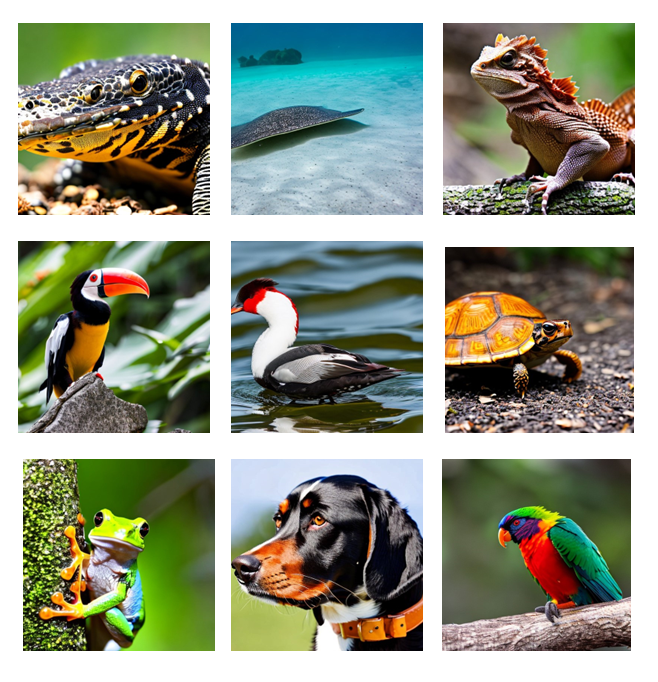}
\caption{RLDF produces photo-realistic ImageNet Clones.}
\label{fig2}
\end{figure}

\section{Related Work}
\label{sec:prior}

Of the rapidly improving photorealistic text-to-image models \cite{a19, a20, a21,a22,a23,a24} we select Stable Diffusion \cite{a19} in favor of reproducibility and demonstrate the model-agnostic capabilities across both open and closed models in the experiments. \\ \textbf{ Guided diffusion models} Fine-tuning and guidance have been shown to be useful for personalized generation \cite{a11,a12,a13} in diffusion models. Depending on the target task, these signals can be set, or the models can be further tuned for specific data generation \cite{a35,a36,a37,a38,a39,a40}.\\ 
\textbf{Prompt Engineering for Diffusion Models}
Prompt-image alignment has been tackled using approaches like gradient-based optimization \cite{a7,a30}, universal guidance \cite{a8}, diffusion process inversion \cite{a9} and extends to applications in video generation \cite{a7}.\\ 
\textbf{Synthetic Data from Diffusion Models} Augmenting real world data with synthetic examples \cite{a27,a28, marathe2023wedge} has been of interest, consequently extending to diffusion models, both fine-tuned and off-the-shelf \cite{a25,a26, a31}. We create similar ImageNet clones with RLDF for comparison in this work.\\ 
\textbf{Diffusion Models and Reinforcement Learning} Previous works have explored RL for guiding diffusion processes \cite{a1},  applied generative modeling for decision-making \cite{a2,a3, a4, a5} and enhanced fine-tuning \cite{a32,a6,a34}. Specifically, for prompt-image alignment, relevant closer work in reinforcement learning includes decision-making frameworks \cite{a6}, prompt optimization \cite{a29,a30,a31}.  \\ We learn through rewards extracted from diffusion generations for image-prompt alignment and do not perform objective-driven diffusion model training or data augmentation.  Additionally, we eliminate the need for human prompting (and guidance) or text input by starting generation from random noise and directly learning from visual semantics. 
\\ Among previous works \cite{qnew,qnew1,qnew2,qnew3} on other tasks, Q* search \cite{qnew}, is a search algorithm showcasing great scalability and experimental results on problems including the Rubik’s cube. 
\\
\textbf{Memorization in Diffusion Models }Due to the volume and varied sources of training data, certain concerning patterns were observed to have emerged in diffusion models. Diffusion models tend to replicate training examples \cite{a14,a15,a16}, which may contain copyright material or artistic styles. Recent work has demonstrated the treatment of this problem as ablation \cite{a17,a18}. We demonstrate the effectiveness of our proposed method RLDF in ablating input concepts like artistic styles and copyrighted characters in this work.

\section{Methodology}
\label{sec:method}

In this section, we provide a brief background on the RL problem formulation. The RLDF presentation is heavily inspired from classical RL theory \cite{g1,g2,g3, watkins1989learning,watkins1992q} and borrows the standard learning objectives to create a novel model for semantic-guided image search. 
\begin{figure}[t]
    \centering
	\begin{minipage}[t]{1\linewidth}
		\centering
		\includegraphics[width=1\linewidth]{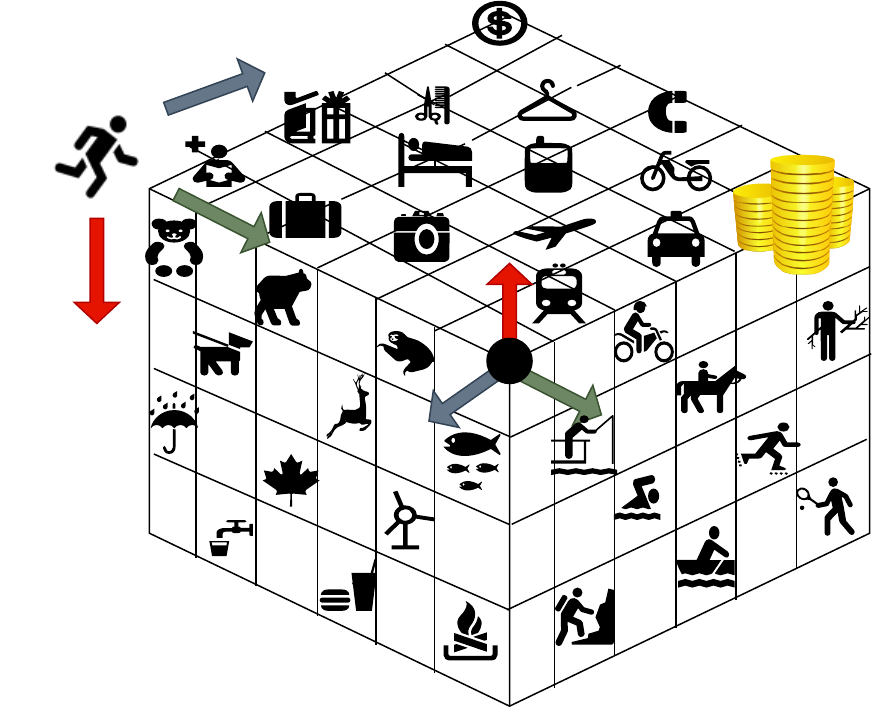} 
	\end{minipage}
	\caption{\label{fig3} \small{RLDF Encoding Space with Semantic Locality}}

\end{figure}

\begin{figure*}[t]
    \centering
	\begin{minipage}[t]{1\linewidth}
		\centering
		\includegraphics[width=1\linewidth]{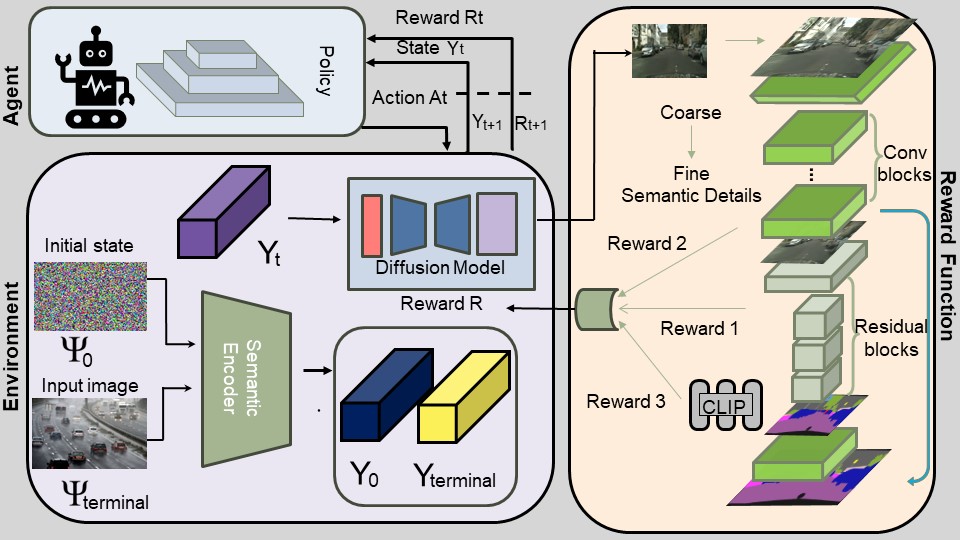} 
	\end{minipage}
	\caption{\label{fig4} \small{RLDF Architecture: The agent receives the environment's encoded state, $\Upsilon_{t} \in \mathcal{\Upsilon}$ at every time step $t$, and selects an action, $A_{t} \in \mathcal{A}(\upsilon) $ .}}

\end{figure*}

\begin{tcolorbox}

\textbf{Formulation: Imagine an n-dimensional gridworld problem where the agent navigates through the object-action space under reinforcement learning policies incentivized by diffusion feedback and semantic guidance.
 }

\end{tcolorbox}

\paragraph{}

  \begin{definition}
    A countable (finite) MDP can be defined by a tuple $\langle\mathcal{\Upsilon}, \mathcal{A}, P, R, \gamma\rangle$ where $\mathcal{\Upsilon}$ is the ``state space'', $\mathcal{A}$ is the ``action space'', $P : \mathcal{\Upsilon} \times \mathcal{A} \to \Delta(\mathcal{\Upsilon})$ is a ``transition kernel'', R is the reward function and $\gamma \in [0, 1)$ is the ``discount factor''.
    \label{def}
  \end{definition}

In RLDF, we formulate the image search problem as an MDP \ref{def} with the classic RL objective \cite{g1} of maximizing the reward $R$ as shown in Figure \ref{fig4}. Over a discrete time step sequence $t=0,1,2,3,..$ we simulate interaction between the diffusion environment and the actor efficiently. The environment initialization places the agent at a random noise encoding state. The agent receives the environment's encoded state, $\Upsilon_{t} \in \mathcal{\Upsilon}$ at every time step $t$, and selects an action, $A_{t} \in \mathcal{A}(\upsilon)$ from this knowledge of the state. After the passing of a time step, the agent obtains a numerical semantic reward, $R_{t+1} \in \mathcal{R} \subset \mathbb{R}$, and is now transported to the new encoded state, $\Upsilon_{t+1} \cdot$ This interaction between the agent and diffusion feedback is well understood as a trajectory: $
\Upsilon_{0}, A_{0}, R_{1}, \Upsilon_{1}, A_{1}, R_{2}, \Upsilon_{2}, A_{2}, R_{3}, \ldots
$. The RLDF trajectory can be visualized as a sequence of images as shown in Figure \ref{fig12}. In each step, the encoding state is the basis of the diffusion model generation as a semantic representation equivalent to a positional state. The dynamics of this system are captured in probability by $p: \mathcal{\Upsilon} \times \mathcal{R} \times \mathcal{\Upsilon} \times \mathcal{A} \rightarrow[0,1]$ which is a deterministic function computing a probability for the values for $R_{t}$ and $\Upsilon_{t}$, dependent only on the preceding state and action. 
\begin{equation}
\resizebox{.9\hsize}{!}{$p\left(\upsilon^{\prime}, r \mid \upsilon, a\right) \doteq    \operatorname{Pr}\left\{\Upsilon_{t}=\upsilon^{\prime}, R_{t}=r \mid \Upsilon_{t-1}=\upsilon, A_{t-1}=a\right\}$  }
\end{equation}
\textbf{Semantic Encoding} In RLDF, we propose the representation of a state as a encoded state  derived from Context Free Grammar rules written for image search.  Classically, a Grammar that generates a language L is given by:
$$ 
G=\langle  T, N, S, R\rangle
$$
{\footnotesize

where $T$ , $N$ ,  $S$ ,  $R$,  $X$, $W$  are the sets of  terminals, non-terminals, start symbol, rules of the form $X \rightarrow W$, non-terminals and a sequence of terminals and non-terminals and: \\  \\  
$N=\{\mathrm{S}, \mathrm{P}, \mathrm{NP}, \mathrm{A}, \mathrm{DP}, \mathrm{PM},\mathrm{I}, \mathrm{LC}, \mathrm{H}, \mathrm{C}, \mathrm{F},  \mathrm{S}, \mathrm{NOM}, \\
\mathrm{VP}$, Noun, Verb, Frequency, Density, Scene$\}$ \\ 
$T = \{$ vocabulary of objects, verbs, actions, scenes, conjunctions, numbers $\}$ \\ 
$S=S$\\
$R= \\ 
\{$ 

$\mathrm{S} \rightarrow $\quad$ $P  $\mathrm{NP}$  A $ \mathrm{DP}$ I  $\mathrm{LC} \quad$ P $\rightarrow$  $\{$ a photo of $\}$     

$\mathrm{NP} \rightarrow $  Frequency NOM  Frequency $\rightarrow$ one 
$\mid$ many $\mid$ ...  

$\mathrm{DP} \rightarrow$ Density PM  Density $\rightarrow$ no 
$\mid$ one $\mid$ ... 

$\mathrm{NOM} \rightarrow $ Noun  Noun $\rightarrow$ banana 
$\mid$ monkey $\mid$ dog $\mid$ ...  

$\mathrm{LC} \rightarrow $ Scene $\quad$ Scene $\rightarrow$  farm 
$\mid$ playground $\mid$  ...  

$\mathrm{VP} \rightarrow$ Verb  $\quad$ Verb $\rightarrow$  playing 
$\mid$ teaching $\mid$   ...  

$\mathrm{PM} \rightarrow$ H  $\quad$ H $\rightarrow$ people 

$\mathrm{A} \rightarrow$ F  $\quad$ F $\rightarrow$ and 

$\mathrm{I} \rightarrow$ C  $\quad$ C $\rightarrow$ in  \\ 
\}
}

\textbf{Intuition} The raw state $\psi$ derived from the above grammar rules is then compressed into a single vector which contains the semantic elements label-encoded by mapping onto the structure of natural language. This represents the encoded state $\upsilon$ which maps the semantics of an entire image  as a vector in the object-action-scene space. Thus, each point in the encoded space can represent one or more images  (many-to-one mapping) that contain the same semantic information. Intuitively, our goal state or input image (gold coins (see Figure \ref{fig3}), is the end point of a reward-guided path. This can be conceptualized as semantic goal-conditional RL, as the target image semantics are used in reward computation. The entire environment is in the encoding space, and each axis represents an image property, thus each action (moving forward, up etc) leads to a new state produced by the diffusion model. However, due to the reward seeking behavior one does not need to stop at the goal, but can continue training with the same objective to seek more rewards and simulate finer semantic attributes.

\textbf{Semantic Locality} The axes representing image properties are the backbone of the control enabled output that this model generates. From both computational and interpretability perspectives, we design the vocabulary tree such that semantic properties that are similar visually are also closely located in the encoding space. Taking the ``scene or background'' visual property as an example, the encodings of park and vegetable garden scenes are closer than the encodings of park and train station platform. This is useful when we use RLDF for precise control(Figure \ref{fig13}).

\begin{algorithm}
\caption{Reinforcement Learning from Diffusion Feedback}
\begin{algorithmic}[1]

\Procedure{RLDF}{image $\omega$}     
    \State Encode $\omega$ $\rightarrow$ terminal 
    \State Initialize $Q(\upsilon, a)$, for all $\upsilon \in \mathcal{\Upsilon}^{+}, a \in \mathcal{A}(\upsilon)$ randomly 
    \State Set Reward Function and Hyperparameter $\epsilon$ \Comment{Sampling from 3 Rewards}
    \While{ episodes not over}
        \State Initialize $\Upsilon$
        \While{ $\Upsilon$ not terminal}
            \State Take action $A$, observe $R, \Upsilon^{\prime}$ from the diffusion model's generated image
            \If{$Q-Learning$}
                \State  $Q(\Upsilon, A) \leftarrow Q(\Upsilon, A) +\alpha\left[R+\gamma \max _{a} Q\left(\Upsilon^{\prime}, a\right) -Q(\Upsilon, A)\right] $                
                \State $\Upsilon \leftarrow \Upsilon^{\prime}$
            \ElsIf{$SARSA$}
                \State Choose $A^{\prime}$ from $\Upsilon^{\prime}$ using policy derived from $Q$
                \State $Q(\Upsilon, A) \leftarrow Q(\Upsilon, A)+\alpha\left[R+\gamma Q\left(\Upsilon^{\prime}, A^{\prime}\right)-Q(\Upsilon, A)\right]$
                \State $\Upsilon \leftarrow \Upsilon^{\prime} $ 
                \State $A \leftarrow A^{\prime}$
            \Else
                \State Random Selection
                \State $\Upsilon \leftarrow \Upsilon^{\prime} $ 
                \State $A \leftarrow A^{\prime}$
            \EndIf
        \EndWhile   
    \EndWhile

\EndProcedure

\end{algorithmic}
\end{algorithm}

\begin{figure}[t]
    \centering
		\centering
		\includegraphics[width=1\linewidth]{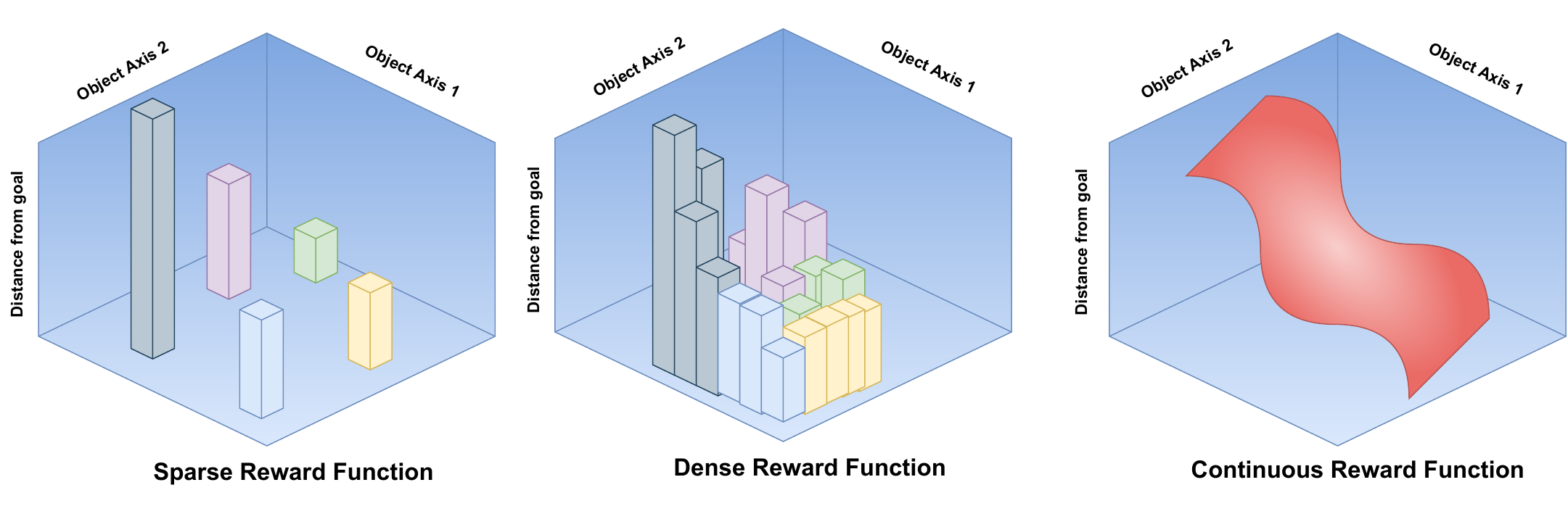} 

	\caption{\label{fig5} \small{The proposed Reward Functions 1, 2 and 3 can roughly be categorized as dense, sparse and semi-continuous.}}

\end{figure}

\textbf{Reward Seeking Behavior} A strong estimation of the gains of a given encoded state or given action is the ideal outcome of this learning paradigm. The state-value function of a state $\upsilon$ under a policy $\pi$ (policy  is a state $\rightarrow$ action selection probability mapping), denoted $v_{\pi}(\upsilon)$, is the expected return for following $\pi$ from $\upsilon$. Expanding on this using a recursive relation: 

$$
\begin{aligned}
v_{\pi}(\upsilon) & \doteq \mathbb{E}_{\pi}\left[G_{t} \mid \Upsilon_{t}=\upsilon\right] \\
&=\sum_{a} \pi(a \mid \upsilon) \sum_{\upsilon^{\prime}, r} p\left(\upsilon^{\prime}, r \mid \upsilon, a\right)\left[r+\gamma v_{\pi}\left(\upsilon^{\prime}\right)\right], \\
& \text { for all } \upsilon \in \mathcal{\Upsilon}
\end{aligned}
$$

The  corresponding action-value function from following a policy $\pi$, denoted $q_{\pi}(\upsilon, a)$, as the expected return for $a$ from $\upsilon$, is given by:

$$
\begin{aligned}
q_{\pi}(\upsilon, a) \doteq \mathbb{E}_{\pi}\left[G_{t} \mid \Upsilon_{t}=\upsilon, A_{t}=a\right] \\ 
=\mathbb{E}_{\pi}\left[\sum_{k=0}^{\infty} \gamma^{k} R_{t+k+1} \mid \Upsilon_{t}=\upsilon, A_{t}=a\right]
\end{aligned}
$$

\textbf{Reward Engineering} We propose 3 reward functions (See Figure \ref{fig5}) which vary in their respective mechanisms of computing semantic gains (``diffusion feedback''). The first reward is ``Multi-Semantic Reward'' and returns high rewards for matching semantic elements between generations with ground truth.
\begin{equation}
    Reward (R) =  \mathbf{1}_{\{g\in G\}} + \mathbf{1}_{\{s\}}
\end{equation}
\begin{equation} 
\mathbf{1}_{\{g\in G\}} = \begin{cases}
C & \text{if } g \in GT[objects],\\
0 & \text{if } g \notin GT[objects[.
\end{cases}
\end{equation}
\begin{equation}
\mathbf{1}_{\{s\}} = \begin{cases}
+C_s & \text{if } g \in GT[scenes],\\
-C_s & \text{if } g \notin GT[scenes].
\end{cases}
\end{equation}

 The second reward is ``Partial-Semantic Reward'' and returns high rewards for matching scene semantics between generations with ground truth.
\begin{equation}
    Reward (R) =  \mathbf{1}_{\{s\}}  = \begin{cases}
+C_s & \text{if } g \in GT[scenes],\\
-C_s & \text{if } g \notin GT[scenes].
\end{cases}
\end{equation}

 The third reward is ``CLIP Reward'' \cite{radford2021learning}  and returns CLIP embedding similarity as a reward between generations with ground truth. Here x and y are the CLIP feature embeddings of the ground truth image and the generation.
\begin{equation}
    Reward (R) =  Cosine(x,y) = \frac{x \cdot y}{|x||y|}
\end{equation}

\textbf{Bellman Optimality} Under an optimal policy $\pi_*$, the expected return for the best action from that state must be equal to the value. The optimality equations for $v_*$ and  $q_*$ are given by: \\  \\ 
{\small
$v_*(\upsilon)  =$ 
$$
\begin{aligned}
\max _{a \in \mathcal{A}(\upsilon)} q_{\pi_*}(\upsilon, a) =\max _a \mathbb{E}_{\pi_*}[G_t \mid \Upsilon_t=\upsilon, A_t=a] \\ 
=\max _a \sum_{\upsilon^{\prime}, r} p\left(\upsilon^{\prime}, r \mid \upsilon, a\right)\left[r+\gamma v_*\left(\upsilon^{\prime}\right)\right] .
\end{aligned}
$$

where the expected discounted return is given by
$G_t$, not to be confused with the grammar notation G.

$q_*(\upsilon, a) $
$$
\begin{aligned}
=\mathbb{E}[R_{t+1}+\gamma \max _{a^{\prime}} q_*(\Upsilon_{t+1}, a^{\prime}) \mid \Upsilon_t=\upsilon  , A_t=a] \\ 
=\sum_{\upsilon^{\prime}, r} p(\upsilon^{\prime}, r \mid \upsilon, a) 
 [r+\gamma \max _{a^{\prime}} q_*(\upsilon^{\prime}, a^{\prime})]
\end{aligned}
$$
}

\textbf{Convergence} Under the conditions of a bounded deterministic reward and infinite exploration horizon in our discounted ($\gamma$) finite MDP, we can obtain convergence guarantees \cite{g2} on RLDF using proven theorems. The Q-learning variant in RLDF given by the update rule with step size $\alpha_t\left(\upsilon,a\right)$:
{\normalsize
$$
\begin{aligned}
Q_{t+1}(\upsilon_t, a_t)=Q_t(\upsilon_t, a_t)+\alpha_t(\upsilon_t,  a_t)[r_t+ \gamma \\ 
 \max _{b \in \mathcal{A}} Q_t(\upsilon_{t+1}, b)-Q_t(\upsilon_t, a_t)]
\end{aligned}
$$
}

converges almost surely to the optimal Q-function for all $(\upsilon, a) \in \mathcal{\Upsilon} \times \mathcal{A}$ under the conditions
$$
\sum_t \alpha_t(\upsilon, a)=\infty \quad \sum_t \alpha_t^2(\upsilon, a)<\infty
$$

*The Upper Cased Functions (eg. Q) are the corresponding array estimates. $\mathcal{\Upsilon}^{+}$ is the set of all states, including the terminal state.

\subsection{Implementation details}

\textbf{Dataset Size} For cloning ImageNet \cite{russakovsky2015imagenet, in1, in2}, we use examples (both published in this paper and used in RLDF) from the popular subset (ImageNet Large Scale Visual Recognition Challenge (ILSVRC) 2012-2017 image classification and localization dataset) \cite{in1}. This contains 1000 object classes and respective train-val-test splits with 1,281,167, 50,000 and 100,000 images. We use RLDF for cloning ImageNet, and build our clone at a similar scale for fair comparison. We reproduce the dataset by generating synthetic images for 1000 object classes with train-val-test splits with 1,508,000, 56,000 and 100,000 images. 

\textbf{Vision-language models} Stable Diffusion \cite{a19} is a latent text-to-image diffusion model that was used in this work for the photo-realistic generations and reproducibility. For the key experiments that include the ImageNet clone, we use the HuggingFace Stable Diffusion v1.4 \cite{v1} model. For the extended model-agnostic plug-and-play experiments, we use the Stable Diffusion v2.1 \cite{v2} model from HuggingFace and DALLE-2 (model version on the day of access)  \cite{a20}.

\textbf{Image Generation} We inference  on a distributed multi-GPU (approximately 5 A100s) setup that takes approximately 7 days for ImageNet cloning. For efficiency, we ran inference with float16, with the better DPMSolverMultistepScheduler scheduler \cite{lu2022dpm,lu2022dpm1}, 20 inference steps, attention slicing, and the latest (on access date) autoencoder \cite{v4}. The images were of high quality, generated at 512 × 512 pixels. Along with the positive CFG, we also prompt models with a large list of negative prompts \cite{huggingfaceStabilityaistablediffusionNegative} for better image quality.

\textbf{Reward Functions} For the three proposed rewards functions we compute the coarse semantic reward using representations learned from \cite{zhou2017places} and fine semantic reward from smaller entity recognition. Our implementation is based on different architectures and data sets, including \cite{p1,p2,p3,p4,p5, he2016deep}. The CLIP rewards \cite{radford2021learning} were calculated based on open source \cite{p6,p7}.

\textbf{RLDF} The basic setup of the environment was based on the GridWorld problem given in the textbook \cite{g1} and subsequent open source implementations \cite{githubGitHubMichaeltinsleyGridworldwithQLearningReinforcementLearning}. We claim that this method is highly efficient due to the mitigation of diffusion-model finetuning and focused search guided by semantic rewards. 

The model cost can be computed as (diffusion model inference cost per step + nominal action-encoding modification cost per step + reward computation cost per step) * number of steps). For environments as small as 480 states, this can be as low as 100 steps of training for baseline (high-level) encodings. For ImageNet cloning we restrict the environment to searching along the object axis, but extend to (count + class) multi-object, action, scene and (count + class) people-specific axes for the remaining results. For diversity in the inference step, we increase the control axes vastly and include object and person attributes like location (city), weathers, times-of-day, artistic styles, colors, ages, race (for diversity and inclusion of humans from all across the world), emotions, and gender axes. The goal is to create well-balanced images that can be used for model training in future work. 

\textbf{Evaluation} We evaluate the RLDF datasets with two key experiments:
\begin{enumerate}
    \item \textbf{Classification}: We trained ResNet-18 \cite{he2016deep}  on Synthetic RLDF ImageNet-100 data and tested on the real ImageNet-100 data \cite{in1}, using  standard classes \cite{githubGitHubDanielchyehImageNet100Pytorch} and training recipes. We also tested on synthetic data from a previous baseline \cite{a26} for an extended comparison.
    \item \textbf{Image Distributions} We compare the RLDF synthetic data with ImageNet \cite{in1} and its natural distribution shifted datasets including ImageNet-Sketch \cite{k1}, ImageNet-R \cite{k2}, ImageNet-A \cite{k3}, ImageNet-O \cite{k3} and compare with results from a previous baseline \cite{a26}. The FID scores \cite{heusel2017gans} and KID scores \cite{binkowski2018demystifying} were implemented using \cite{parmar2021cleanfid}.
\end{enumerate}

\textbf{Methodology} The design of the context-free grammar encoding was inspired by early work \cite{e1,e2}. In this work we use also data samples or reference diagram elements from open sources \cite{pathakICLR18zeroshot, kenk2020dawn, russakovsky2015imagenet, radford2021learning, wang2018pix2pixHD,kang2023gigagan}.

\textbf{Applications} Of the many interesting generative applications, we focus on \cite{a26} for previous ImageNet experiments, \cite{a17} for demonstrating use of RLDF for similar ablation of memorized concepts and \cite{anew} for using RLDF for similar precise control over attributes by sliding over axes of the RLDF environment.

\textbf{Noisy Diffusion Gradient} While formulating the RLDF MDP, the intermediate outputs are semantic encodings. By directly computing the gradients ($\Delta f$) on these encodings, we propose a fast noisy diffusion gradient model to reach the global optima (input image encoding). This does not have similar guarantees on convergence due to the noise from the diffusion model, and often gets stuck in plateaus.  Noise here means the possibility of diffusion models generating artifacts harming semantic registration. The key disadvantages of this proposed method are the failure modes under low signals and fast divergence under post-goal training conditions.

\begin{figure}
\includegraphics[width = 1\linewidth]{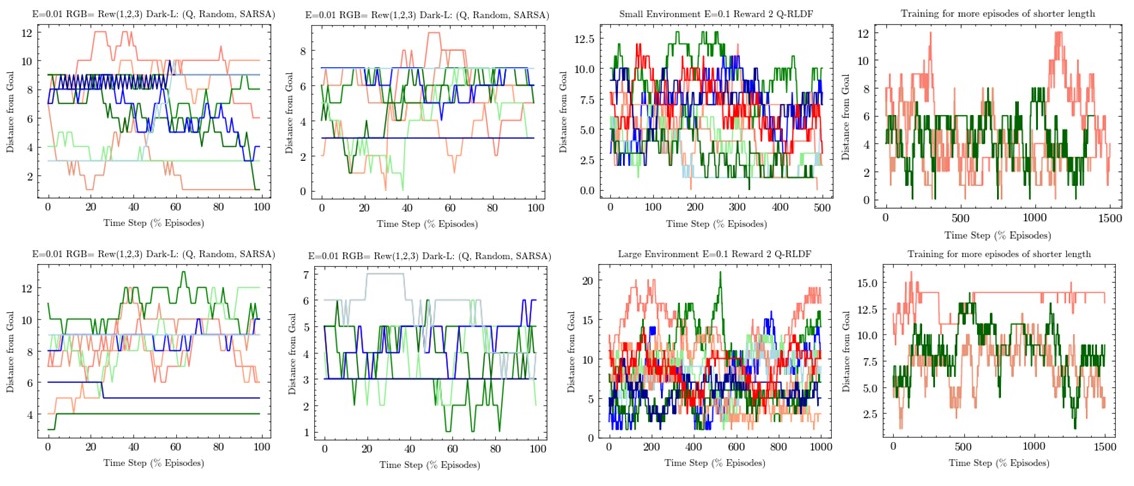}
\caption{Trajectory of agents marked by distance from terminal state. Using $\epsilon=0.1$ and Partial-Semantic Reward, the agent can maximize reward seeking behavior over longer trajectories and has more stable training patterns as well. }
\label{fig6}
\end{figure}

\begin{figure}
\includegraphics[width = 1\linewidth]{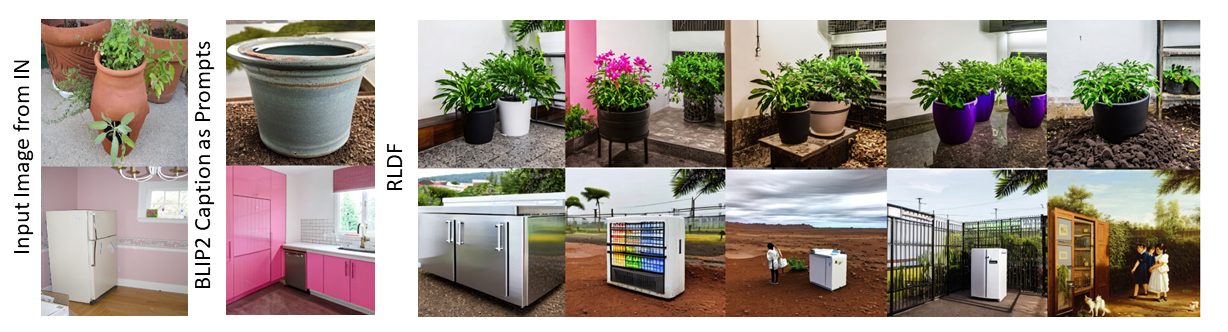}
\caption{\textbf{Why RLDF? Just feed captions as diffusion prompts instead!} Image captioning (Column 2) is evaluated by metrics that might not directly translate to diffusion output success. We show in BLIP2 captions \cite{blip}  may not necessarily provide the diffusion model with the needed semantic information.}
\label{fig7}
\end{figure}

\begin{figure}[h]
\includegraphics[width = 1\linewidth]{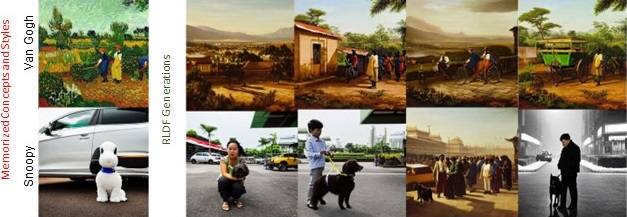}
\caption{\textbf{Previous work \cite{a17} show that certain styles and concepts are memorized exactly and can be removed using their proposed method. We similarly show that RLDF can effectively (and implicitly) erase memorized concepts like Van Gogh artistic style or Snoopy character and replace them with generic styles instead. }}
\label{fig8}
\end{figure}

\begin{figure}[h]
\includegraphics[width = 1\linewidth]{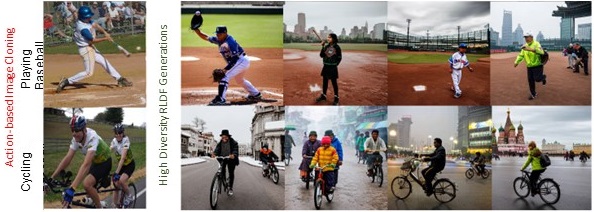}
\caption{\textbf{As most of the previous experiments highlight RLDF in the object space, here we showcase examples where RLDF can generalize in the action space to actions like playing baseball and cycling and learns the semantics respectively.}}
\label{fig9}
\end{figure}

\begin{figure}[h]
\includegraphics[width = 1\linewidth]{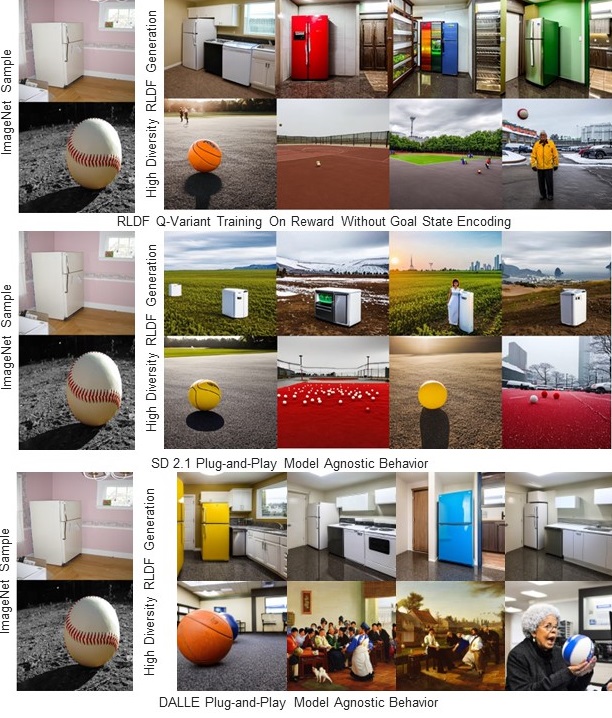}
\caption{\textbf{We demonstrate how RLDF continues training with the reward-seeking objective and can keep generating desirable images. The model-agnostic nature of this model allows us to plug-and-play any text-to-image models like SD 2.1 or DALLE-2. The images in this figure are generated using RLDF and SD models only for reproducibility (the DALLE-2 section was trained on DALLE-2 feedback and inferenced using SD for reproducibility}). }
\label{fig10}
\end{figure}

\begin{figure}[h]
\includegraphics[width = 1\linewidth]{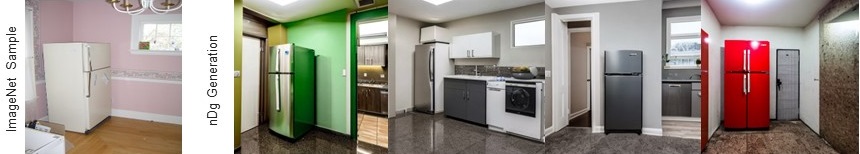}
\caption{\textbf{The optimization driven Noisy Diffusion Gradient method is able to reach the target encoding and generate images satisfactorily while retaining semantic information.}}
\label{fig11}
\end{figure}

\begin{figure}[h]
\includegraphics[width = 1\linewidth]{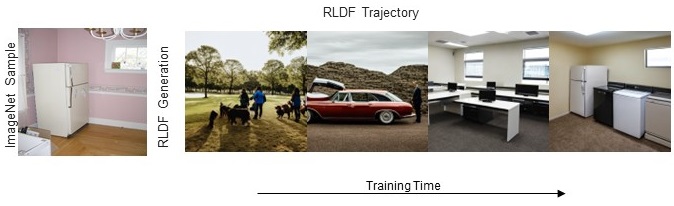}
\caption{\textbf{The RLDF trajectory can be visualized as a sequence of images corresponding to the respective semantic encodings. As we train the models, we get closer to the desired image. For visualization, we skip the intermediate steps and only show model results at significant intermediate intervals. }}
\label{fig12}
\end{figure}

\begin{figure}[h]
\includegraphics[width = 1\linewidth]{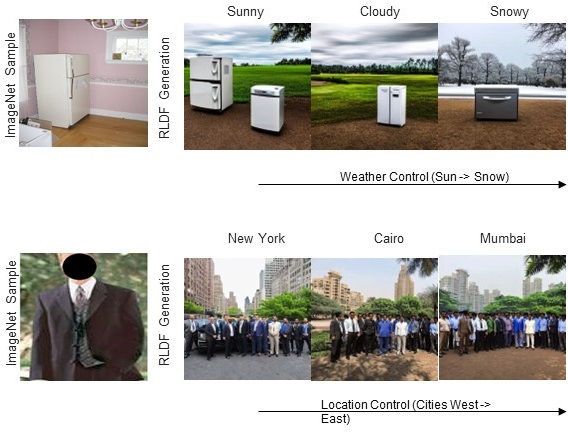}
\caption{\textbf{Previous works \cite{anew} propose Concept Sliders for Precise Control. We apply RLDF for this application and are able to observe control over weather and location attributes. It is important to note that we do not memorize input objects and focus on class-consistency over object-consistency and also impute bias-equalizing noise for diversity, thus leading to different looking refrigerators or crowds.}}
\label{fig13}
\end{figure}

\begin{figure}[h]
\centering
\includegraphics[width = 1\linewidth]{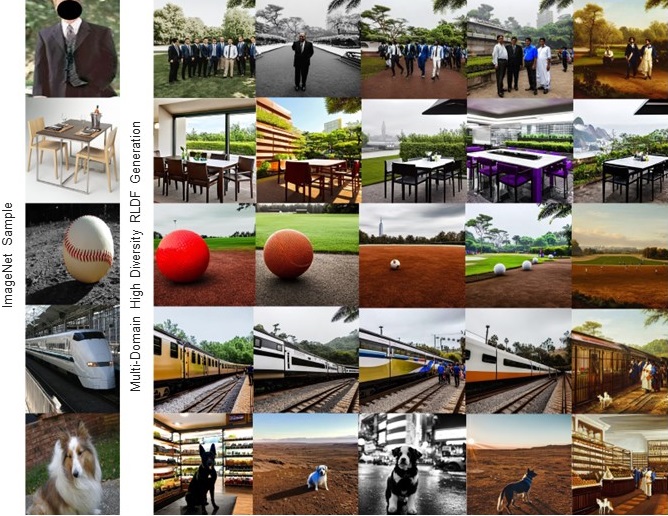}
\caption{\textbf{We test over multiple domain inputs like clothing, retail, sports, transport, agriculture, botany, zoology and obtain highly realistic yet semantically similar outputs as shown with the examples in the above figure.}}
\label{fig14}
\end{figure}

\begin{table}[]
\begin{tabular}{llllll}
\hline
\textbf{Data} & \textbf{IN} & \textbf{Sketch} & \textbf{Im-R}   & \textbf{Im-A} & \textbf{Im-O} \\ \hline
Real  \cite{a26}        & NR              & 248                & 225    & NR            & NR            \\
Gen \cite{a26}    & NR              & 210                & 190    & NR            & NR            \\
RLDF    & \cellcolor{tabfirst}180.8          & \cellcolor{tabfirst}255.3             & \cellcolor{tabfirst}227.5 & \cellcolor{tabfirst}303.7        & \cellcolor{tabfirst} 264.1        \\
KID    & 0.08            & 0.1              & 0.13   & 0.18          & 0.06          \\ \hline
\end{tabular}
\caption{RLDF produces ImageNet Clones that are comparatively closer to the real IN in FID, but far from NDS datasets among the compared datasets.}
\label{tab3}
\end{table}

\begin{figure}[h]
\includegraphics[width = 1\linewidth]{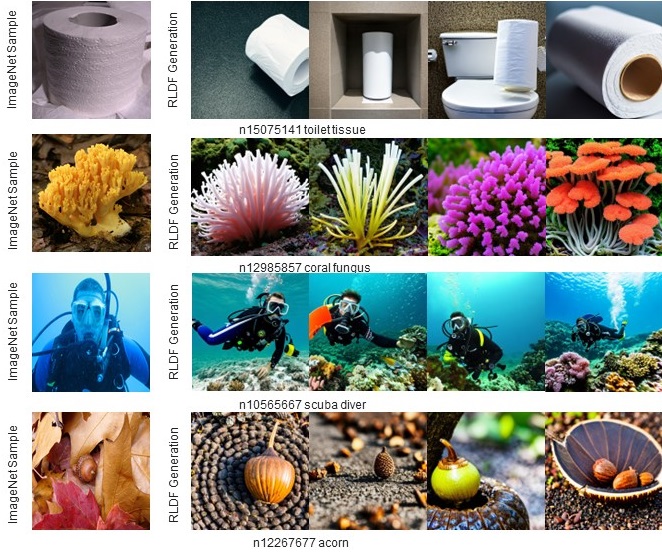}
\caption{\textbf{We obtain photo-realistic ImageNet clones that are faithful to the original classes.}}
\label{fig15}
\end{figure}

\begin{table}
\begin{center}    
\begin{tabular}{ll}
\hline
\textbf{Dataset}                                                                &  Acc@1 \\ \hline
Validation Set (Our clone)                                                    & \cellcolor{tabfirst} 85.11                                                            \\
\begin{tabular}[c]{@{}l@{}}ImageNet100\\ (Random Sample)\end{tabular}  & \cellcolor{tabthird} 12.42                                                              \\
\begin{tabular}[c]{@{}l@{}}ImageNet100\\ (Validation Set)\end{tabular} & \cellcolor{tabsecond} 13.16                                                      \\
Generated IN100 \cite{a26}                                                     & \cellcolor{tabfirst} 49.14                                                 \\ \hline
\end{tabular}

\caption{\label{tab2}ResNet-18 Performance trained only on SD 1.4 RLDF synthetic data.}
\end{center}

\end{table}
\begin{table}[]
\begin{center}
\resizebox{1\columnwidth}{!}{
\begin{tabular}{llllllllllll} \hline
 Agent & Reward & $\epsilon$ & $D_T$ & $D_{max}$ & $D_{min}$ & $\rho$ & $\sigma^2$ & Conv. & F Semantic & C Semantic \\ \hline

Q	&	1	&	\cellcolor{tabfirst}0.01	&	\cellcolor{tabfirst}1.0	&	\cellcolor{tabthird}7.0	&	\cellcolor{tabsecond}1.0	&	1.58	&	2.51	&	\cellcolor{tabsecond}0.0	&	\cellcolor{tabthird}33.0	&	\cellcolor{tabsecond}43.0	\\
Q	&	1	&	\cellcolor{tabsecond}0.10	&	10	&	12	&	8	&	0.97	&	0.95	&	\cellcolor{tabsecond}0.0	&	0	&	0	\\
Q	&	2	&	\cellcolor{tabfirst}0.01	&	\cellcolor{tabfirst}1.0	&	9	&	\cellcolor{tabsecond}1.0	&	1.61	&	2.59	&	\cellcolor{tabsecond}0.0	&	18	&	1	\\
Q	&	2	&	\cellcolor{tabsecond}0.10	&	9	&	9	&	7	&	0.67	&	0.45	&	\cellcolor{tabsecond}0.0	&	0	&	\cellcolor{tabfirst}52.0	\\
Q	&	3	&	\cellcolor{tabfirst}0.01	&	9	&	10	&	7	&	0.55	&	0.3	&	\cellcolor{tabsecond}0.0	&	0	&	0	\\
Q	&	3	&	\cellcolor{tabsecond}0.10	&	\cellcolor{tabsecond}2.0	&	\cellcolor{tabfirst}3.0	&	\cellcolor{tabsecond}1.0	&	0.26	&	0.07	&	\cellcolor{tabsecond}0.0	&	26	&	0	\\
Random	&	1	&	\cellcolor{tabfirst}0.01	&	6	&	12	&	6	&	1.51	&	2.27	&	\cellcolor{tabsecond}0.0	&	0	&	0	\\
Random	&	1	&	\cellcolor{tabsecond}0.10	&	9	&	11	&	\cellcolor{tabfirst}0.0	&	\cellcolor{tabfirst}3.15	&	\cellcolor{tabfirst}9.92	&	\cellcolor{tabfirst}1.0	&	13	&	6	\\
Random	&	2	&	\cellcolor{tabfirst}0.01	&	8	&	9	&	5	&	1.04	&	1.08	&	\cellcolor{tabsecond}0.0	&	0	&	20	\\
Random	&	2	&	\cellcolor{tabsecond}0.10	&	9	&	11	&	6	&	1.05	&	1.09	&	\cellcolor{tabsecond}0.0	&	0	&	11	\\
Random	&	3	&	\cellcolor{tabfirst}0.01	&	4	&	9	&	3	&	1.58	&	2.5	&	\cellcolor{tabsecond}0.0	&	11	&	0	\\
Random	&	3	&	\cellcolor{tabsecond}0.10	&	\cellcolor{tabsecond}2.0	&	\cellcolor{tabsecond}5.0	&	\cellcolor{tabthird}2.0	&	0.79	&	0.63	&	\cellcolor{tabsecond}0.0	&	25	&	0	\\
SARSA	&	1	&	\cellcolor{tabfirst}0.01	&	10	&	10	&	7	&	0.85	&	0.73	&	\cellcolor{tabsecond}0.0	&	13	&	0	\\
SARSA	&	1	&	\cellcolor{tabsecond}0.10	&	\cellcolor{tabfirst}1.0	&	\cellcolor{tabsecond}5.0	&	\cellcolor{tabfirst}0.0	&	1.51	&	2.29	&	\cellcolor{tabfirst}1.0	&	3	&	10	\\
SARSA	&	2	&	\cellcolor{tabfirst}0.01	&	\cellcolor{tabthird}3.0	&	\cellcolor{tabsecond}5.0	&	3	&	0.57	&	0.33	&	\cellcolor{tabsecond}0.0	&	3	&	\cellcolor{tabthird}42.0	\\
SARSA	&	2	&	\cellcolor{tabsecond}0.10	&	9	&	9	&	\cellcolor{tabthird}2.0	&	\cellcolor{tabthird}2.08	&	\cellcolor{tabthird}4.34	&	\cellcolor{tabsecond}0.0	&	\cellcolor{tabsecond}36.0	&	0	\\
SARSA	&	3	&	\cellcolor{tabfirst}0.01	&	9	&	10	&	3	&	\cellcolor{tabsecond}2.84	&	\cellcolor{tabsecond}8.09	&	\cellcolor{tabsecond}0.0	&	0	&	0	\\
SARSA	&	3	&	\cellcolor{tabsecond}0.10	&	9	&	9	&	\cellcolor{tabfirst}0.0	&	1.64	&	2.68	&	\cellcolor{tabfirst}1.0	&	\cellcolor{tabfirst}51.0	&	10	\\  \hline
                 
\end{tabular}}
\caption{\label{tab1}Training Statistics for RLDF:  
 We observe that 0.1 is an optimal $\epsilon$ that balances exploration and exploitation. In this table the parameters are the distance from goal with statistics (maximum, minimum, standard deviation, variance), intermediate goals, fine and coarse semantic gains for the first environment training runs. }
\end{center}

\end{table}

\section{Results}

The goal of RLDF for ImageNet cloning was to imitate natural semantic elements through diffusion feedback learning. We do not aim to improve the image-generation process (e.g. diffusion) itself in any way, and rather focus on obtaining semantically similar images. We find that RLDF is able to overcome previous semantic challenges and generate consistent distributions across ImageNet-1k classes, as visible in Figures \ref{fig2} and \ref{fig15}. 
For this task, we generate 1.5 M images for all 1000 classes,  following the ImageNet \cite{russakovsky2015imagenet} dataset structure and train the classifier using standard recipes. Congruent with findings from previous works, we do find that training classifiers on purely synthetic data does not generalize well to real world performance and present our results for 100 classes (ImageNet-100) standalone in Table \ref{tab2}. Interestingly, in Table \ref{tab3}, we see that using the FID metrics, the RLDF ImageNet Clone lies comparatively closer to original ImageNet distribution and farther from its natural distribution shifts like ImageNet-Sketch \cite{k3,k2,k1} among the compared datasets.

\section{Discussion}

We present some key insights from the RLDF training runs (see Figure \ref{fig6})  including: 
\begin{enumerate}
    \item The training patterns displayed in Table \ref{tab1} heavily depend on the initial position of the agent in the encoding space (as this governs the extent to which the agent can explore in near neighborhood to find terminal state faster) which has been treated through random initialization.
    \item To maintain the reward and state consistency, we are required to perform appropriate seeding which determines the state-reward as generated from the diffusion model. Using dynamic seeds during the training process is not recommended, as the model may produce diverging semantics  giving looping rewards along the same initial trajectory.
    \item CLIP Rewards  tend to provide better fine-grained guidance, however gets stuck at local optima in the reward landscape, which may cause missing the desired location in the exploratory phase of the agent. In contrast, Partial Semantic Guidance provides a more gradual reward landscape, thus ensuring communication between immediate reward and final convergence goal closer to the input image distribution semantically.
    \item 
We intentionally do not include distance from terminal state as a reward due to the ambiguity of Euclidean distance. Two images equally far away from the destination with high semantic differences should ideally have different rewards.
 \item Increasing the world dimensions (we scale the base environment by 9x) can increase compute over 2x, yet the RLDF learner is still able to efficiently locate the desired properties in the encoding space.
 \item The individual task difficulty can often be domain specific. For example, in the indoor navigation domain when a bowl is kept on a table, the random agent was also able to reach the end of the trials. In other domains, RL agents performed significantly better.
 \item In small state space exploration environments, random action sampling($\epsilon =  1$) appears to yield instances of randomly good performance. However, when we scale this for deeper exploration, one finds that the earlier observed performance was spurious as the agent finds themselves much farther away from the ideal state through random sampling versus learning better actions through exploiting seen information. 
 \item Design Idea: In the problem setup, one can penalize the agent for unrealistic generations by setting their encodings as terminal states or negative rewards. We avoid this in our work to isolate the model learning from human feedback and guidance.

\end{enumerate}

\section{Conclusion and Future Work}

We show that RLDF model can produce semantic-rich generations through class-aware guidance and diffusion feedback, showcased in Figure \ref{fig1}. Our second proposed method (Noisy Diffusion Gradient) also obtains semantically accurate results, as shown in Figure \ref{fig11}. The key contribution is efficient compression of semantics into encoded vectors to represent any real world image. \textbf{Critically, RLDF requires no text input, text guidance  or fine-tuning of TTI models.} In the future works, one can plug in better text-to-image models for improving performance on real-world benchmarks.

\textbf{Applications }RLDF generalizes across object (Figure \ref{fig2}) and action (Figure \ref{fig9}) spaces. The plug-and-play mechanism allows for substitution of underlying TTI models (Figure \ref{fig10}). We show that feeding  captions (Figure \ref{fig7}) may not supply sufficient semantic information, which RLDF attempts to remedy.   RLDF can be applied for ablation of memorized concepts (Figure \ref{fig8}) and for similar precise control over attributes (Figure \ref{fig13}) by sliding over axes of the RLDF environment.

\textbf{Limitations} Some limitations of RLDF include computational costs in larger environments, subject inconsistency (we focus on class-consistency instead), and the constraints of being bounded in performance to the goodness of the underlying text-to-image model.

\section{Acknowledgements}

Gratitude to God, my parents, teachers, and friends for providing me with resources, guidance, knowledge and cheesy pasta in this journey. Thanks to the AlphaGo documentary, which kindled my interest in reinforcement learning. This work is but a small sapling of my ideas, with the desire to grow and blossom with feedback (human/diffusion).

\clearpage
{
    \small
    \bibliographystyle{ieeenat_fullname}
    \bibliography{main}

\begin{thebibliography}{83}
\providecommand{\natexlab}[1]{#1}
\providecommand{\url}[1]{\texttt{#1}}
\expandafter\ifx\csname urlstyle\endcsname\relax
  \providecommand{\doi}[1]{doi: #1}\else
  \providecommand{\doi}{doi: \begingroup \urlstyle{rm}\Url}\fi

\bibitem[git({\natexlab{a}})]{githubGitHubDanielchyehImageNet100Pytorch}
{G}it{H}ub - danielchyeh/{I}mage{N}et-100-{P}ytorch: ({P}ytorch) {T}raining {R}es{N}ets on {I}mage{N}et-100 data --- github.com.
\newblock \url{https://github.com/danielchyeh/ImageNet-100-Pytorch}, {\natexlab{a}}.
\newblock [Accessed 23-11-2023].

\bibitem[git({\natexlab{b}})]{githubGitHubMichaeltinsleyGridworldwithQLearningReinforcementLearning}
{G}it{H}ub - michaeltinsley/{G}ridworld-with-{Q}-{L}earning-{R}einforcement-{L}earning-: {J}upyter notebook containing a solution to {S}utton and {B}arto's gridworld problem with both a random agent and a {Q}-learning agent. --- github.com.
\newblock \url{https://github.com/michaeltinsley/Gridworld-with-Q-Learning-Reinforcement-Learning-}, {\natexlab{b}}.
\newblock [Accessed 23-11-2023].

\bibitem[hug()]{huggingfaceStabilityaistablediffusionNegative}
stabilityai/stable-diffusion · {N}egative {P}rompts --- huggingface.co.
\newblock \url{https://huggingface.co/spaces/stabilityai/stable-diffusion/discussions/7857}.
\newblock [Accessed 23-11-2023].

\bibitem[p5()]{p5}
{G}it{H}ub - nathen418/{V}ehicle-{T}racking-{U}sing-{O}pen{C}{V}-and-{V}{O}{L}{O}v5: {I}n {D}evelopment - {A} vehicle tracker written in {P}ython using {O}pen{C}{V} and {Y}{O}{L}{O}v5 -- ({V}{T}{U}{O}{V}) --- github.com.
\newblock \url{https://github.com/nathen418/Vehicle-Tracking-Using-OpenCV-and-VOLOv5/tree/main}.
\newblock [Accessed 23-11-2023].

\bibitem[p6()]{p6}
openai/clip-vit-large-patch14 · {H}ugging {F}ace --- huggingface.co.
\newblock \url{https://huggingface.co/openai/clip-vit-large-patch14}.
\newblock [Accessed 23-11-2023].

\bibitem[p7()]{p7}
{G}it{H}ub - openai/{C}{L}{I}{P}: {C}{L}{I}{P} ({C}ontrastive {L}anguage-{I}mage {P}retraining), {P}redict the most relevant text snippet given an image --- github.com.
\newblock \url{https://github.com/openai/CLIP}.
\newblock [Accessed 23-11-2023].

\bibitem[v1()]{v1}
{C}omp{V}is/stable-diffusion-v1-4 · {H}ugging {F}ace --- huggingface.co.
\newblock \url{https://huggingface.co/CompVis/stable-diffusion-v1-4}.
\newblock [Accessed 23-11-2023].

\bibitem[v2()]{v2}
stabilityai/stable-diffusion-2-1 · {H}ugging {F}ace --- huggingface.co.
\newblock \url{https://huggingface.co/stabilityai/stable-diffusion-2-1}.
\newblock [Accessed 23-11-2023].

\bibitem[v4()]{v4}
stabilityai/sd-vae-ft-mse-original · {H}ugging {F}ace --- huggingface.co.
\newblock \url{https://huggingface.co/stabilityai/sd-vae-ft-mse-original}.
\newblock [Accessed 23-11-2023].

\bibitem[Addison~Howard(2018)]{in1}
Wendy~Kan Addison~Howard, Eunbyung~Park.
\newblock Imagenet object localization challenge, 2018.

\bibitem[Agostinelli et~al.(2021)Agostinelli, Shmakov, McAleer, Fox, and Baldi]{qnew}
Forest Agostinelli, Alexander Shmakov, Stephen McAleer, Roy Fox, and Pierre Baldi.
\newblock A* search without expansions: Learning heuristic functions with deep q-networks.
\newblock \emph{arXiv preprint arXiv:2102.04518}, 2021.

\bibitem[Ajay et~al.(2022)Ajay, Du, Gupta, Tenenbaum, Jaakkola, and Agrawal]{a2}
Anurag Ajay, Yilun Du, Abhi Gupta, Joshua Tenenbaum, Tommi Jaakkola, and Pulkit Agrawal.
\newblock Is conditional generative modeling all you need for decision-making?
\newblock \emph{arXiv preprint arXiv:2211.15657}, 2022.

\bibitem[Anonymous(2023{\natexlab{a}})]{a4}
Anonymous.
\newblock Diffusion world models.
\newblock In \emph{Submitted to The Twelfth International Conference on Learning Representations}, 2023{\natexlab{a}}.
\newblock under review.

\bibitem[Anonymous(2023{\natexlab{b}})]{a7}
Anonymous.
\newblock Text-aware diffusion policies.
\newblock In \emph{Submitted to The Twelfth International Conference on Learning Representations}, 2023{\natexlab{b}}.
\newblock under review.

\bibitem[Anonymous(2023{\natexlab{c}})]{a8}
Anonymous.
\newblock {DPO}-diff: On discrete prompt optimization of text-to-image diffusion models.
\newblock In \emph{Submitted to The Twelfth International Conference on Learning Representations}, 2023{\natexlab{c}}.
\newblock under review.

\bibitem[Anonymous(2023{\natexlab{d}})]{a9}
Anonymous.
\newblock Reverse stable diffusion: What prompt was used to generate this image?
\newblock In \emph{Submitted to The Twelfth International Conference on Learning Representations}, 2023{\natexlab{d}}.
\newblock under review.

\bibitem[Azizi et~al.(2023)Azizi, Kornblith, Saharia, Norouzi, and Fleet]{a27}
Shekoofeh Azizi, Simon Kornblith, Chitwan Saharia, Mohammad Norouzi, and David~J Fleet.
\newblock Synthetic data from diffusion models improves imagenet classification.
\newblock \emph{arXiv preprint arXiv:2304.08466}, 2023.

\bibitem[Ball et~al.(2023)Ball, Smith, Kostrikov, and Levine]{qnew2}
Philip~J Ball, Laura Smith, Ilya Kostrikov, and Sergey Levine.
\newblock Efficient online reinforcement learning with offline data.
\newblock \emph{arXiv preprint arXiv:2302.02948}, 2023.

\bibitem[Bansal and Grover(2023)]{a26}
Hritik Bansal and Aditya Grover.
\newblock Leaving reality to imagination: Robust classification via generated datasets.
\newblock \emph{arXiv preprint arXiv:2302.02503}, 2023.

\bibitem[Bellman(1957)]{g3}
Richard Bellman.
\newblock A markovian decision process.
\newblock \emph{Journal of mathematics and mechanics}, pages 679--684, 1957.

\bibitem[Bi{\'n}kowski et~al.(2018)Bi{\'n}kowski, Sutherland, Arbel, and Gretton]{binkowski2018demystifying}
Miko{\l}aj Bi{\'n}kowski, Danica~J Sutherland, Michael Arbel, and Arthur Gretton.
\newblock Demystifying mmd gans.
\newblock \emph{arXiv preprint arXiv:1801.01401}, 2018.

\bibitem[Black et~al.(2023)Black, Janner, Du, Kostrikov, and Levine]{a6}
Kevin Black, Michael Janner, Yilun Du, Ilya Kostrikov, and Sergey Levine.
\newblock Training diffusion models with reinforcement learning.
\newblock \emph{arXiv preprint arXiv:2305.13301}, 2023.

\bibitem[Carlini et~al.(2023)Carlini, Hayes, Nasr, Jagielski, Sehwag, Tramer, Balle, Ippolito, and Wallace]{a16}
Nicolas Carlini, Jamie Hayes, Milad Nasr, Matthew Jagielski, Vikash Sehwag, Florian Tramer, Borja Balle, Daphne Ippolito, and Eric Wallace.
\newblock Extracting training data from diffusion models.
\newblock In \emph{32nd USENIX Security Symposium (USENIX Security 23)}, pages 5253--5270, 2023.

\bibitem[Chang et~al.(2023)Chang, Zhang, Barber, Maschinot, Lezama, Jiang, Yang, Murphy, Freeman, Rubinstein, et~al.]{a24}
Huiwen Chang, Han Zhang, Jarred Barber, AJ Maschinot, Jose Lezama, Lu Jiang, Ming-Hsuan Yang, Kevin Murphy, William~T Freeman, Michael Rubinstein, et~al.
\newblock Muse: Text-to-image generation via masked generative transformers.
\newblock \emph{arXiv preprint arXiv:2301.00704}, 2023.

\bibitem[Chebotar et~al.(2023)Chebotar, Vuong, Irpan, Hausman, Xia, Lu, Kumar, Yu, Herzog, Pertsch, et~al.]{qnew1}
Yevgen Chebotar, Quan Vuong, Alex Irpan, Karol Hausman, Fei Xia, Yao Lu, Aviral Kumar, Tianhe Yu, Alexander Herzog, Karl Pertsch, et~al.
\newblock Q-transformer: Scalable offline reinforcement learning via autoregressive q-functions.
\newblock \emph{arXiv preprint arXiv:2309.10150}, 2023.

\bibitem[Chen et~al.(2023)Chen, Sun, Song, and Luo]{a36}
Shoufa Chen, Peize Sun, Yibing Song, and Ping Luo.
\newblock Diffusiondet: Diffusion model for object detection.
\newblock In \emph{Proceedings of the IEEE/CVF International Conference on Computer Vision}, pages 19830--19843, 2023.

\bibitem[Chomsky(1956)]{e2}
Noam Chomsky.
\newblock Three models for the description of language.
\newblock \emph{IRE Transactions on information theory}, 2\penalty0 (3):\penalty0 113--124, 1956.

\bibitem[Deng et~al.(2009)Deng, Dong, Socher, Li, Li, and Fei-Fei]{in2}
Jia Deng, Wei Dong, Richard Socher, Li-Jia Li, Kai Li, and Li Fei-Fei.
\newblock Imagenet: A large-scale hierarchical image database.
\newblock In \emph{2009 IEEE Conference on Computer Vision and Pattern Recognition}, pages 248--255, 2009.

\bibitem[Esser et~al.(2023)Esser, Chiu, Atighehchian, Granskog, and Germanidis]{a39}
Patrick Esser, Johnathan Chiu, Parmida Atighehchian, Jonathan Granskog, and Anastasis Germanidis.
\newblock Structure and content-guided video synthesis with diffusion models.
\newblock In \emph{Proceedings of the IEEE/CVF International Conference on Computer Vision}, pages 7346--7356, 2023.

\bibitem[Fan et~al.(2023)Fan, Watkins, Du, Liu, Ryu, Boutilier, Abbeel, Ghavamzadeh, Lee, and Lee]{a32}
Ying Fan, Olivia Watkins, Yuqing Du, Hao Liu, Moonkyung Ryu, Craig Boutilier, Pieter Abbeel, Mohammad Ghavamzadeh, Kangwook Lee, and Kimin Lee.
\newblock Reinforcement learning for fine-tuning text-to-image diffusion models.
\newblock In \emph{Thirty-seventh Conference on Neural Information Processing Systems}, 2023.

\bibitem[Gal et~al.(2022)Gal, Alaluf, Atzmon, Patashnik, Bermano, Chechik, and Cohen-Or]{a13}
Rinon Gal, Yuval Alaluf, Yuval Atzmon, Or Patashnik, Amit~H Bermano, Gal Chechik, and Daniel Cohen-Or.
\newblock An image is worth one word: Personalizing text-to-image generation using textual inversion.
\newblock \emph{arXiv preprint arXiv:2208.01618}, 2022.

\bibitem[Gandikota et~al.(2023)Gandikota, Materzynska, Zhou, Torralba, and Bau]{anew}
Rohit Gandikota, Joanna Materzynska, Tingrui Zhou, Antonio Torralba, and David Bau.
\newblock Concept sliders: Lora adaptors for precise control in diffusion models, 2023.

\bibitem[Golatkar et~al.(2020)Golatkar, Achille, and Soatto]{a18}
Aditya Golatkar, Alessandro Achille, and Stefano Soatto.
\newblock Eternal sunshine of the spotless net: Selective forgetting in deep networks.
\newblock In \emph{Proceedings of the IEEE/CVF Conference on Computer Vision and Pattern Recognition}, pages 9304--9312, 2020.

\bibitem[Hao et~al.(2022)Hao, Chi, Dong, and Wei]{a29}
Yaru Hao, Zewen Chi, Li Dong, and Furu Wei.
\newblock Optimizing prompts for text-to-image generation.
\newblock \emph{arXiv preprint arXiv:2212.09611}, 2022.

\bibitem[Hassan()]{p1}
Zubair Hassan.
\newblock {C}ar {D}etection using {O}pen{C}{V} and {P}ython within 5 minutes! - {F}olio3{A}{I} {B}log --- folio3.ai.
\newblock \url{https://www.folio3.ai/blog/car-detection-using-opencv-and-python-within-5-minutes/}.
\newblock [Accessed 23-11-2023].

\bibitem[He et~al.(2016)He, Zhang, Ren, and Sun]{he2016deep}
Kaiming He, Xiangyu Zhang, Shaoqing Ren, and Jian Sun.
\newblock Deep residual learning for image recognition.
\newblock In \emph{Proceedings of the IEEE conference on computer vision and pattern recognition}, pages 770--778, 2016.

\bibitem[He et~al.(2022)He, Sun, Yu, Xue, Zhang, Torr, Bai, and Qi]{a28}
Ruifei He, Shuyang Sun, Xin Yu, Chuhui Xue, Wenqing Zhang, Philip Torr, Song Bai, and Xiaojuan Qi.
\newblock Is synthetic data from generative models ready for image recognition?
\newblock \emph{arXiv preprint arXiv:2210.07574}, 2022.

\bibitem[Hendrycks et~al.(2021{\natexlab{a}})Hendrycks, Basart, Mu, Kadavath, Wang, Dorundo, Desai, Zhu, Parajuli, Guo, Song, Steinhardt, and Gilmer]{k2}
Dan Hendrycks, Steven Basart, Norman Mu, Saurav Kadavath, Frank Wang, Evan Dorundo, Rahul Desai, Tyler Zhu, Samyak Parajuli, Mike Guo, Dawn Song, Jacob Steinhardt, and Justin Gilmer.
\newblock The many faces of robustness: A critical analysis of out-of-distribution generalization.
\newblock \emph{ICCV}, 2021{\natexlab{a}}.

\bibitem[Hendrycks et~al.(2021{\natexlab{b}})Hendrycks, Zhao, Basart, Steinhardt, and Song]{k3}
Dan Hendrycks, Kevin Zhao, Steven Basart, Jacob Steinhardt, and Dawn Song.
\newblock Natural adversarial examples.
\newblock \emph{CVPR}, 2021{\natexlab{b}}.

\bibitem[Heusel et~al.(2017)Heusel, Ramsauer, Unterthiner, Nessler, and Hochreiter]{heusel2017gans}
Martin Heusel, Hubert Ramsauer, Thomas Unterthiner, Bernhard Nessler, and Sepp Hochreiter.
\newblock Gans trained by a two time-scale update rule converge to a local nash equilibrium.
\newblock \emph{Advances in neural information processing systems}, 30, 2017.

\bibitem[Hopcroft et~al.(2006)Hopcroft, Motwani, and Ullman]{e1}
John~E Hopcroft, Rajeev Motwani, and Jeffrey~D Ullman.
\newblock Automata theory, languages, and computation.
\newblock \emph{International Edition}, 24\penalty0 (2):\penalty0 171--183, 2006.

\bibitem[Jain and Ravanbakhsh(2023)]{a5}
Vineet Jain and Siamak Ravanbakhsh.
\newblock Learning to reach goals via diffusion.
\newblock \emph{arXiv preprint arXiv:2310.02505}, 2023.

\bibitem[Janner et~al.(2022)Janner, Du, Tenenbaum, and Levine]{a1}
Michael Janner, Yilun Du, Joshua Tenenbaum, and Sergey Levine.
\newblock Planning with diffusion for flexible behavior synthesis.
\newblock In \emph{International Conference on Machine Learning}, pages 9902--9915. PMLR, 2022.

\bibitem[Kang et~al.(2023)Kang, Zhu, Zhang, Park, Shechtman, Paris, and Park]{kang2023gigagan}
Minguk Kang, Jun-Yan Zhu, Richard Zhang, Jaesik Park, Eli Shechtman, Sylvain Paris, and Taesung Park.
\newblock Scaling up gans for text-to-image synthesis.
\newblock In \emph{Proceedings of the IEEE Conference on Computer Vision and Pattern Recognition (CVPR)}, 2023.

\bibitem[Kawar et~al.(2023)Kawar, Zada, Lang, Tov, Chang, Dekel, Mosseri, and Irani]{a37}
Bahjat Kawar, Shiran Zada, Oran Lang, Omer Tov, Huiwen Chang, Tali Dekel, Inbar Mosseri, and Michal Irani.
\newblock Imagic: Text-based real image editing with diffusion models.
\newblock In \emph{Proceedings of the IEEE/CVF Conference on Computer Vision and Pattern Recognition}, pages 6007--6017, 2023.

\bibitem[Kenk and Hassaballah(2020)]{kenk2020dawn}
Mourad~A Kenk and Mahmoud Hassaballah.
\newblock Dawn: vehicle detection in adverse weather nature dataset.
\newblock \emph{arXiv preprint arXiv:2008.05402}, 2020.

\bibitem[Kostrikov et~al.(2021)Kostrikov, Nair, and Levine]{qnew3}
Ilya Kostrikov, Ashvin Nair, and Sergey Levine.
\newblock Offline reinforcement learning with implicit q-learning.
\newblock \emph{arXiv preprint arXiv:2110.06169}, 2021.

\bibitem[Kumari et~al.(2023)Kumari, Zhang, Wang, Shechtman, Zhang, and Zhu]{a17}
Nupur Kumari, Bingliang Zhang, Sheng-Yu Wang, Eli Shechtman, Richard Zhang, and Jun-Yan Zhu.
\newblock Ablating concepts in text-to-image diffusion models.
\newblock In \emph{Proceedings of the IEEE/CVF International Conference on Computer Vision}, pages 22691--22702, 2023.

\bibitem[Li et~al.(2023{\natexlab{a}})Li, Prabhudesai, Duggal, Brown, and Pathak]{a31}
Alexander~C Li, Mihir Prabhudesai, Shivam Duggal, Ellis Brown, and Deepak Pathak.
\newblock Your diffusion model is secretly a zero-shot classifier.
\newblock \emph{arXiv preprint arXiv:2303.16203}, 2023{\natexlab{a}}.

\bibitem[Li et~al.(2023{\natexlab{b}})Li, Li, Savarese, and Hoi]{blip}
Junnan Li, Dongxu Li, Silvio Savarese, and Steven Hoi.
\newblock Blip-2: Bootstrapping language-image pre-training with frozen image encoders and large language models.
\newblock \emph{arXiv preprint arXiv:2301.12597}, 2023{\natexlab{b}}.

\bibitem[Lin et~al.(2023)Lin, Gao, Tang, Takikawa, Zeng, Huang, Kreis, Fidler, Liu, and Lin]{a35}
Chen-Hsuan Lin, Jun Gao, Luming Tang, Towaki Takikawa, Xiaohui Zeng, Xun Huang, Karsten Kreis, Sanja Fidler, Ming-Yu Liu, and Tsung-Yi Lin.
\newblock Magic3d: High-resolution text-to-3d content creation.
\newblock In \emph{Proceedings of the IEEE/CVF Conference on Computer Vision and Pattern Recognition}, pages 300--309, 2023.

\bibitem[Lin et~al.(2014)Lin, Maire, Belongie, Hays, Perona, Ramanan, Doll{\'a}r, and Zitnick]{p3}
Tsung-Yi Lin, Michael Maire, Serge Belongie, James Hays, Pietro Perona, Deva Ramanan, Piotr Doll{\'a}r, and C~Lawrence Zitnick.
\newblock Microsoft coco: Common objects in context.
\newblock In \emph{Computer Vision--ECCV 2014: 13th European Conference, Zurich, Switzerland, September 6-12, 2014, Proceedings, Part V 13}, pages 740--755. Springer, 2014.

\bibitem[Lu et~al.(2022{\natexlab{a}})Lu, Zhou, Bao, Chen, Li, and Zhu]{lu2022dpm}
Cheng Lu, Yuhao Zhou, Fan Bao, Jianfei Chen, Chongxuan Li, and Jun Zhu.
\newblock Dpm-solver: A fast ode solver for diffusion probabilistic model sampling in around 10 steps.
\newblock \emph{Advances in Neural Information Processing Systems}, 35:\penalty0 5775--5787, 2022{\natexlab{a}}.

\bibitem[Lu et~al.(2022{\natexlab{b}})Lu, Zhou, Bao, Chen, Li, and Zhu]{lu2022dpm1}
Cheng Lu, Yuhao Zhou, Fan Bao, Jianfei Chen, Chongxuan Li, and Jun Zhu.
\newblock Dpm-solver++: Fast solver for guided sampling of diffusion probabilistic models.
\newblock \emph{arXiv preprint arXiv:2211.01095}, 2022{\natexlab{b}}.

\bibitem[Marathe et~al.(2023)Marathe, Ramanan, Walambe, and Kotecha]{marathe2023wedge}
Aboli Marathe, Deva Ramanan, Rahee Walambe, and Ketan Kotecha.
\newblock Wedge: A multi-weather autonomous driving dataset built from generative vision-language models.
\newblock In \emph{Proceedings of the IEEE/CVF Conference on Computer Vision and Pattern Recognition}, pages 3317--3326, 2023.

\bibitem[Melo()]{g2}
Francisco~S Melo.
\newblock Convergence of q-learning: a simple proof.

\bibitem[Mokady et~al.(2023)Mokady, Hertz, Aberman, Pritch, and Cohen-Or]{a40}
Ron Mokady, Amir Hertz, Kfir Aberman, Yael Pritch, and Daniel Cohen-Or.
\newblock Null-text inversion for editing real images using guided diffusion models.
\newblock In \emph{Proceedings of the IEEE/CVF Conference on Computer Vision and Pattern Recognition}, pages 6038--6047, 2023.

\bibitem[Nichol et~al.(2021)Nichol, Dhariwal, Ramesh, Shyam, Mishkin, McGrew, Sutskever, and Chen]{a21}
Alex Nichol, Prafulla Dhariwal, Aditya Ramesh, Pranav Shyam, Pamela Mishkin, Bob McGrew, Ilya Sutskever, and Mark Chen.
\newblock Glide: Towards photorealistic image generation and editing with text-guided diffusion models.
\newblock \emph{arXiv preprint arXiv:2112.10741}, 2021.

\bibitem[Parmar et~al.(2022)Parmar, Zhang, and Zhu]{parmar2021cleanfid}
Gaurav Parmar, Richard Zhang, and Jun-Yan Zhu.
\newblock On aliased resizing and surprising subtleties in gan evaluation.
\newblock In \emph{CVPR}, 2022.

\bibitem[Pathak et~al.(2018)Pathak, Mahmoudieh, Luo, Agrawal, Chen, Shentu, Shelhamer, Malik, Efros, and Darrell]{pathakICLR18zeroshot}
Deepak Pathak, Parsa Mahmoudieh, Guanghao Luo, Pulkit Agrawal, Dian Chen, Yide Shentu, Evan Shelhamer, Jitendra Malik, Alexei~A. Efros, and Trevor Darrell.
\newblock Zero-shot visual imitation.
\newblock In \emph{ICLR}, 2018.

\bibitem[Prabhudesai et~al.(2023)Prabhudesai, Goyal, Pathak, and Fragkiadaki]{a30}
Mihir Prabhudesai, Anirudh Goyal, Deepak Pathak, and Katerina Fragkiadaki.
\newblock Aligning text-to-image diffusion models with reward backpropagation.
\newblock \emph{arXiv preprint arXiv:2310.03739}, 2023.

\bibitem[Radford et~al.(2021)Radford, Kim, Hallacy, Ramesh, Goh, Agarwal, Sastry, Askell, Mishkin, Clark, et~al.]{radford2021learning}
Alec Radford, Jong~Wook Kim, Chris Hallacy, Aditya Ramesh, Gabriel Goh, Sandhini Agarwal, Girish Sastry, Amanda Askell, Pamela Mishkin, Jack Clark, et~al.
\newblock Learning transferable visual models from natural language supervision.
\newblock In \emph{International conference on machine learning}, pages 8748--8763. PMLR, 2021.

\bibitem[Ramesh et~al.(2022)Ramesh, Dhariwal, Nichol, Chu, and Chen]{a20}
Aditya Ramesh, Prafulla Dhariwal, Alex Nichol, Casey Chu, and Mark Chen.
\newblock Hierarchical text-conditional image generation with clip latents.
\newblock \emph{arXiv preprint arXiv:2204.06125}, 1\penalty0 (2):\penalty0 3, 2022.

\bibitem[Redmon and Farhadi(2018)]{p2}
Joseph Redmon and Ali Farhadi.
\newblock Yolov3: An incremental improvement.
\newblock \emph{arXiv preprint arXiv:1804.02767}, 2018.

\bibitem[Rombach et~al.(2022)Rombach, Blattmann, Lorenz, Esser, and Ommer]{a19}
Robin Rombach, Andreas Blattmann, Dominik Lorenz, Patrick Esser, and Bj{\"o}rn Ommer.
\newblock High-resolution image synthesis with latent diffusion models.
\newblock In \emph{Proceedings of the IEEE/CVF conference on computer vision and pattern recognition}, pages 10684--10695, 2022.

\bibitem[Ruiz et~al.(2023)Ruiz, Li, Jampani, Pritch, Rubinstein, and Aberman]{a11}
Nataniel Ruiz, Yuanzhen Li, Varun Jampani, Yael Pritch, Michael Rubinstein, and Kfir Aberman.
\newblock Dreambooth: Fine tuning text-to-image diffusion models for subject-driven generation.
\newblock In \emph{Proceedings of the IEEE/CVF Conference on Computer Vision and Pattern Recognition}, pages 22500--22510, 2023.

\bibitem[Russakovsky et~al.(2015)Russakovsky, Deng, Su, Krause, Satheesh, Ma, Huang, Karpathy, Khosla, Bernstein, et~al.]{russakovsky2015imagenet}
Olga Russakovsky, Jia Deng, Hao Su, Jonathan Krause, Sanjeev Satheesh, Sean Ma, Zhiheng Huang, Andrej Karpathy, Aditya Khosla, Michael Bernstein, et~al.
\newblock Imagenet large scale visual recognition challenge.
\newblock \emph{International journal of computer vision}, 115:\penalty0 211--252, 2015.

\bibitem[Saharia et~al.(2022)Saharia, Chan, Saxena, Li, Whang, Denton, Ghasemipour, Gontijo~Lopes, Karagol~Ayan, Salimans, et~al.]{a22}
Chitwan Saharia, William Chan, Saurabh Saxena, Lala Li, Jay Whang, Emily~L Denton, Kamyar Ghasemipour, Raphael Gontijo~Lopes, Burcu Karagol~Ayan, Tim Salimans, et~al.
\newblock Photorealistic text-to-image diffusion models with deep language understanding.
\newblock \emph{Advances in Neural Information Processing Systems}, 35:\penalty0 36479--36494, 2022.

\bibitem[Sar{\i}y{\i}ld{\i}z et~al.(2023)Sar{\i}y{\i}ld{\i}z, Alahari, Larlus, and Kalantidis]{a25}
Mert~B{\"u}lent Sar{\i}y{\i}ld{\i}z, Karteek Alahari, Diane Larlus, and Yannis Kalantidis.
\newblock Fake it till you make it: Learning transferable representations from synthetic imagenet clones.
\newblock In \emph{Proceedings of the IEEE/CVF Conference on Computer Vision and Pattern Recognition}, pages 8011--8021, 2023.

\bibitem[Somepalli et~al.(2023{\natexlab{a}})Somepalli, Singla, Goldblum, Geiping, and Goldstein]{a14}
Gowthami Somepalli, Vasu Singla, Micah Goldblum, Jonas Geiping, and Tom Goldstein.
\newblock Understanding and mitigating copying in diffusion models.
\newblock \emph{arXiv preprint arXiv:2305.20086}, 2023{\natexlab{a}}.

\bibitem[Somepalli et~al.(2023{\natexlab{b}})Somepalli, Singla, Goldblum, Geiping, and Goldstein]{a15}
Gowthami Somepalli, Vasu Singla, Micah Goldblum, Jonas Geiping, and Tom Goldstein.
\newblock Diffusion art or digital forgery? investigating data replication in diffusion models.
\newblock In \emph{Proceedings of the IEEE/CVF Conference on Computer Vision and Pattern Recognition}, pages 6048--6058, 2023{\natexlab{b}}.

\bibitem[Sutton and Barto(2018)]{g1}
Richard~S Sutton and Andrew~G Barto.
\newblock \emph{Reinforcement learning: An introduction}.
\newblock MIT press, 2018.

\bibitem[Team()]{p4}
TechVidvan Team.
\newblock {V}ehicle {C}ounting, {C}lassification \& {D}etection using {O}pen{C}{V} \& {P}ython - {T}ech{V}idvan --- techvidvan.com.
\newblock \url{https://techvidvan.com/tutorials/opencv-vehicle-detection-classification-counting/}.
\newblock [Accessed 23-11-2023].

\bibitem[Wang et~al.(2019)Wang, Ge, Lipton, and Xing]{k1}
Haohan Wang, Songwei Ge, Zachary Lipton, and Eric~P Xing.
\newblock Learning robust global representations by penalizing local predictive power.
\newblock In \emph{Advances in Neural Information Processing Systems}, pages 10506--10518, 2019.

\bibitem[Wang et~al.(2018)Wang, Liu, Zhu, Tao, Kautz, and Catanzaro]{wang2018pix2pixHD}
Ting-Chun Wang, Ming-Yu Liu, Jun-Yan Zhu, Andrew Tao, Jan Kautz, and Bryan Catanzaro.
\newblock High-resolution image synthesis and semantic manipulation with conditional gans.
\newblock In \emph{Proceedings of the IEEE Conference on Computer Vision and Pattern Recognition}, 2018.

\bibitem[Wang et~al.(2022)Wang, Hunt, and Zhou]{a3}
Zhendong Wang, Jonathan~J Hunt, and Mingyuan Zhou.
\newblock Diffusion policies as an expressive policy class for offline reinforcement learning.
\newblock \emph{arXiv preprint arXiv:2208.06193}, 2022.

\bibitem[Watkins and Dayan(1992)]{watkins1992q}
Christopher~JCH Watkins and Peter Dayan.
\newblock Q-learning.
\newblock \emph{Machine learning}, 8:\penalty0 279--292, 1992.

\bibitem[Watkins(1989)]{watkins1989learning}
Christopher John Cornish~Hellaby Watkins.
\newblock Learning from delayed rewards.
\newblock 1989.

\bibitem[Wu et~al.(2023)Wu, Ge, Wang, Lei, Gu, Shi, Hsu, Shan, Qie, and Shou]{a38}
Jay~Zhangjie Wu, Yixiao Ge, Xintao Wang, Stan~Weixian Lei, Yuchao Gu, Yufei Shi, Wynne Hsu, Ying Shan, Xiaohu Qie, and Mike~Zheng Shou.
\newblock Tune-a-video: One-shot tuning of image diffusion models for text-to-video generation.
\newblock In \emph{Proceedings of the IEEE/CVF International Conference on Computer Vision}, pages 7623--7633, 2023.

\bibitem[Xu et~al.(2023)Xu, Liu, Wu, Tong, Li, Ding, Tang, and Dong]{a34}
Jiazheng Xu, Xiao Liu, Yuchen Wu, Yuxuan Tong, Qinkai Li, Ming Ding, Jie Tang, and Yuxiao Dong.
\newblock Imagereward: Learning and evaluating human preferences for text-to-image generation.
\newblock \emph{arXiv preprint arXiv:2304.05977}, 2023.

\bibitem[Yu et~al.(2022)Yu, Xu, Koh, Luong, Baid, Wang, Vasudevan, Ku, Yang, Ayan, et~al.]{a23}
Jiahui Yu, Yuanzhong Xu, Jing~Yu Koh, Thang Luong, Gunjan Baid, Zirui Wang, Vijay Vasudevan, Alexander Ku, Yinfei Yang, Burcu~Karagol Ayan, et~al.
\newblock Scaling autoregressive models for content-rich text-to-image generation.
\newblock \emph{arXiv preprint arXiv:2206.10789}, 2022.

\bibitem[Zhang et~al.(2023)Zhang, Rao, and Agrawala]{a12}
Lvmin Zhang, Anyi Rao, and Maneesh Agrawala.
\newblock Adding conditional control to text-to-image diffusion models.
\newblock In \emph{Proceedings of the IEEE/CVF International Conference on Computer Vision}, pages 3836--3847, 2023.

\bibitem[Zhou et~al.(2017)Zhou, Lapedriza, Khosla, Oliva, and Torralba]{zhou2017places}
Bolei Zhou, Agata Lapedriza, Aditya Khosla, Aude Oliva, and Antonio Torralba.
\newblock Places: A 10 million image database for scene recognition.
\newblock \emph{IEEE Transactions on Pattern Analysis and Machine Intelligence}, 2017.

\end{thebibliography}
}

\end{document}